\pdfoutput=1

\documentclass[letterpaper, 10 pt, conference]{ieeeconf}  %

\IEEEoverridecommandlockouts                              %

\newcommand{\etal}{et al. }
\newcommand{\ie}{i.e. }
\newcommand{\eg}{e.g. }

\usepackage{graphicx} %
\usepackage{amsmath} %
\usepackage{amssymb}  %
\usepackage{color, soul}
\usepackage{url}

\renewcommand{\baselinestretch}{0.93}
\usepackage{etoolbox}
\makeatletter
\patchcmd\@combinedblfloats{\box\@outputbox}{\unvbox\@outputbox}{}{\errmessage{\noexpand patch failed}}
\makeatother

\makeatletter
\let\NAT@parse\undefined
\makeatother
\usepackage{hyperref}

\title{\LARGE \bf
Generative Modeling of Multimodal Multi-Human Behavior
}

\author{Boris Ivanovic$^{1}$ 
\hspace{0.5cm} Edward Schmerling$^{2}$ 
\hspace{0.5cm} Karen Leung$^{3}$ 
\hspace{0.5cm} Marco Pavone$^{3}$%
\thanks{*This work was supported by the Office of Naval Research (Grant N00014-17-1-2433) and by the Toyota Research Institute (``TRI"). This article solely reflects the opinions and conclusions of its authors and not ONR, TRI or any other Toyota entity.}%
\thanks{$^{1}$Boris Ivanovic is with the Department of Computer Science,
        Stanford University, Stanford, CA 94305, USA
        {\tt\small \{borisi@cs.stanford.edu\}}}%
\thanks{$^{2}$Edward Schmerling is with the Institute  for Computational and Mathematical Engineering,
        Stanford University, Stanford, CA 94305, USA
        {\tt\small \{schmrlng@stanford.edu\}}}%
\thanks{$^{3}$Karen Leung and Marco Pavone are with the Department of Aeronautics and Astronautics, 
        Stanford University, Stanford, CA 94305, USA
        {\tt\small \{karenl7, pavone\}@stanford.edu}}%
}

\begin{document}

\maketitle
\thispagestyle{empty}
\pagestyle{empty}

\begin{abstract}

This work presents a methodology for modeling and predicting human behavior in settings with $N$ humans interacting in highly multimodal scenarios (\ie where there are many possible highly-distinct futures). A motivating example includes robots interacting with humans in crowded environments, such as self-driving cars operating alongside human-driven vehicles or human-robot collaborative bin packing in a warehouse. Our approach to model human behavior in such uncertain environments is to model humans in the scene as nodes in a graphical model, with edges encoding relationships between them. For each human, we learn a multimodal probability distribution over future actions from a dataset of multi-human interactions. Learning such distributions is made possible by recent advances in the theory of conditional variational autoencoders and deep learning approximations of probabilistic graphical models. Specifically, we learn action distributions conditioned on interaction history, neighboring human behavior, and candidate future agent behavior in order to take into account response dynamics. We demonstrate the performance of such a modeling approach in modeling basketball player trajectories, a highly multimodal, multi-human scenario which serves as a proxy for many robotic applications.

\end{abstract}

\section{INTRODUCTION}

\begin{figure}[t]
  \centering
  \includegraphics[width=0.85\linewidth]{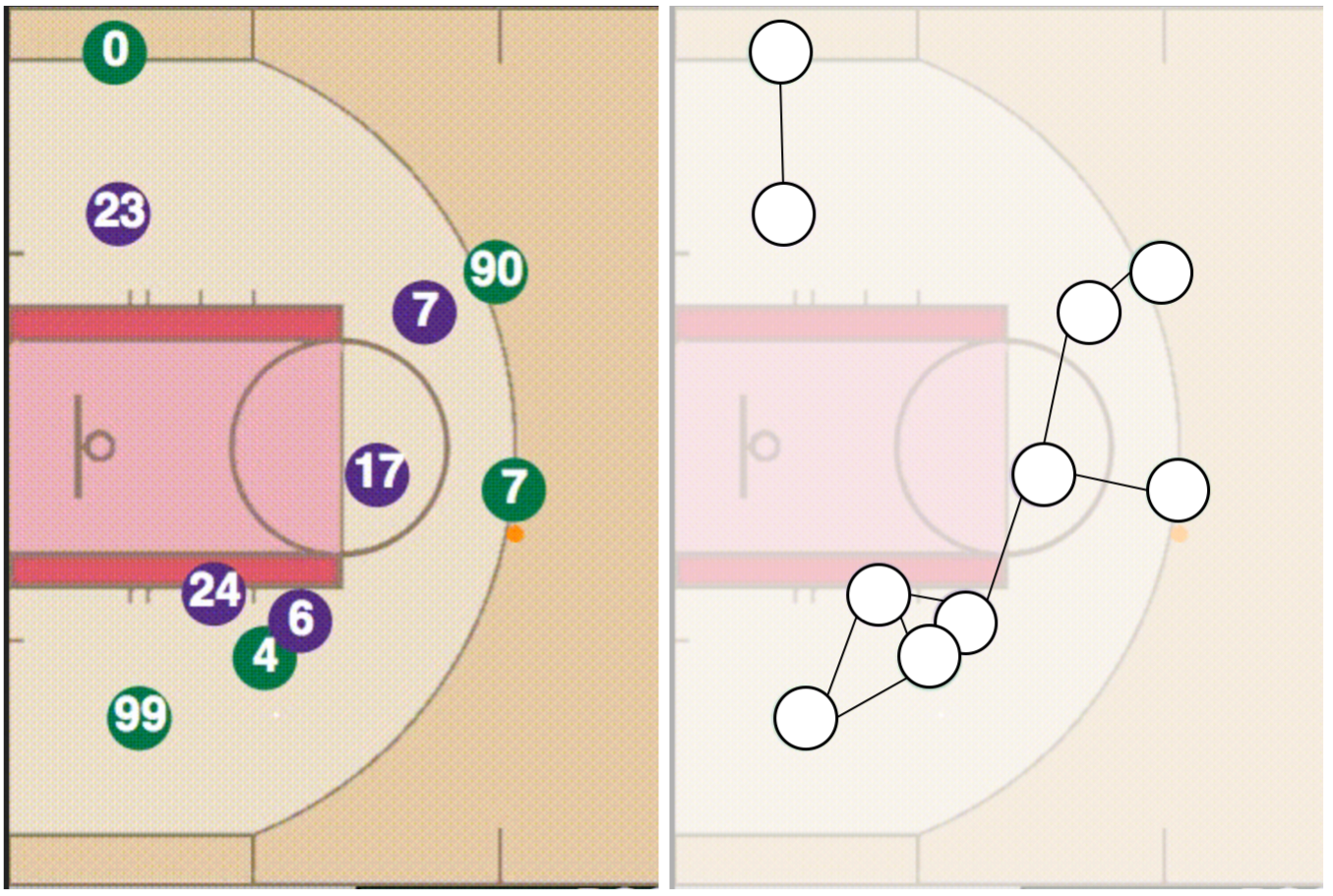}
  \caption{\textbf{Left}: A basketball game from the 2015-16 National Basketball Association (NBA) season. \textbf{Right}: An undirected graph representation of the same scene.}
  \label{fig:scenarios}
\end{figure}

Safe human-robot interaction (HRI) is a necessity before the widespread integration of autonomous systems and society can occur. As a result, most autonomous systems in production today are low-risk systems with minimal human interaction, a fact that will surely change with the 
ever-rising growth of automation in manufacturing, warehouses, and transportation.
However, enabling robots to safely interact and plan with humans in mind comes with a unique set of challenges. One of the most demanding challenges is modeling and predicting the highly multimodal nature of human behavior, where ``multimodal" refers to the possibility of many highly-distinct futures. Complex interaction dynamics found in multi-human environments are too complicated to be simulated and repeatedly generated for learning applications \cite{Handel2016}. This necessitates learning directly from how humans naturally interact in the environment, which limits the rate of data capture, and thus the rate at which learning may occur. Further, it is desirable to perform human-in-the-loop experiments to directly compare decision making performance using such models, which includes considerations of safety and artificial system limitations. This is tolerable in controlled experimental environments, but true autonomy in the uncontrolled world requires the ability to generalize and handle novel situations. To this end, we propose a solution to the following general HRI task: ``Proactively account for the many potential actions of $N$ humans in order to perform a task safely." We emphasize that $N$ is a variable (\ie not constant). We want to learn a model that holds over a range of $N$, not only a chosen value that the model can be tailored to. Being proactive is important as it allows an agent to reason about potential surrounding human behavior, enabling more-informed decisions to be made. We exemplify this general task with basketball playing, a highly multimodal, multi-human interaction scenario (Fig. \ref{fig:scenarios}).

Inspired by \cite{SchmerlingLeungEtAl2018}, we first highlight four important considerations that any desirable modeling framework should address to model $N$ humans in rapidly-evolving multimodal interaction scenarios. First, the framework must be able to model human behavior conditioned on candidate agent actions. This allows for intelligent planning taking into account potential human responses. Second, any modeling framework must be able to handle multimodality since human actions are uncertain and potentially multimodal. Third, the framework must be able to handle longer-term interaction history. This allows for the modeling of higher-level interaction behavior (\ie inferring latent states) which affects the character of future predictions. Finally, any modeling framework must efficiently scale to representing $N$ humans. In the unstructured world, many humans may be interacting with an agent at any time, thus a representation that can seamlessly handle and reason about multi-human interactions is required. We motivate our model design decisions by systematically considering these properties and analyzing prior work in related fields.

{\em Statement of Contribution}:
This paper presents a state-of-the-art data-driven framework for safe HRI that, to the best of our knowledge, uniquely meets all four aforementioned conditions for a desirable modeling framework that models $N$ humans in highly multimodal interaction scenarios. Our approach inherently handles a variable number of humans and does not make any assumptions about their outcome preference or structure of interaction beliefs (\eg as one might see in game theory). Additionally, this modeling framework maintains a degree of interpretability
as we can visualize an agent's belief about human responses to its actions by sampling from our model. We demonstrate its performance in modeling basketball players' trajectories (a highly multimodal, multi-human interaction scenario) and analyze our model's scalability with respect to number of parameters, compute time, and memory usage.

{\em Outline}:
The rest of this paper is structured as follows: In Section \ref{lit_review}, we review prior and related work, while also describing the reasoning that guided our model design decisions. We define our notation and rigorously formulate the problem we wish to address in Section \ref{problem_formulation}. Section \ref{solution} describes the details of our novel approach to model and predict $N$ humans' behavior in highly multimodal scenarios. Empirical evaluation of our model follows in Section \ref{experiments}. Finally, we conclude the paper in Section \ref{conclusion}, where we also provide directions for future work.

\section{RELATED WORK}\label{lit_review}
The first major consideration is the ability to produce multimodal, future-conditional trajectory predictions. This is the problem of predicting a human's many potential actions given an agent's potential future actions. Here, trajectories are sequences of actions and trajectory prediction can be thought of as a regression problem where we wish to predict a human's future actions. Since this is essentially a sequence-modeling problem, three major approaches stand out that have been applied to similar problems and necessitate deeper analysis: Inverse Reinforcement Learning (IRL), Gaussian Process Regression (GPR), and Recurrent Neural Networks (RNNs).

\subsection{Existing Methods}
The problem of trajectory prediction can be formulated as a reward-maximization problem in an IRL framework, and has been done so for American football player movements (rife with challenging interaction scenarios) \cite{LeeKitani2016}. It was shown that a player's path can be modeled as a Markov Decision Process (MDP) where the path reward is learned via maximum entropy IRL on a mixture of static hand-crafted and dynamic features. In the problem formulation of \cite{LeeKitani2016}, Lee and Kitani additionally model an opponent that blocks the agent from completing its goal. This opponent is accounted for in the reward function via a dynamic feature (\ie its predicted position). The opponent's trajectory is learned and predicted with GPR. Specifically, GPR is used to obtain a short-term prediction of the opponent's movement, which is iteratively updated over time. Unfortunately, this opponent model is not multimodal and thus the overall model does not account for multiple opponent behaviors. Their model also relies on predetermined player route plans giving a strong prior on the location and path of the agent, which is unavailable in general settings. A more fundamental limitation is that IRL mostly relies on a unimodal assumption of interaction outcome \cite{KoberBagnellEtAl2013, NgRussell2000}. While this may be remedied with sufficiently complex and numerous features, without any state augmentation the resulting formulation is Markovian and cannot condition on interaction history to reason about the future \cite{NgRussell2000}.

GPR is another strong trajectory prediction framework \cite{RasmussenWilliams2006}. Further, there are formulations which introduce latent variable conditioning to produce different high-level behaviors \cite{RasmussenWilliams2006, WangFleetEtAl2008}. Such works show impressive results, however, the learned algorithms take minutes to hours to run, rendering them infeasible for many robotics use cases. Further, GPR as-is cannot model multimodal data. There have been ensemble methods proposed that handle multimodal data by decomposing the original distribution into smaller parts that are modeled by local GPs (and then aggregated to form one model) \cite{RasmussenWilliams2006, DasSrivastava2010}. However, such approaches similarly fall prey to long inference times, which increase as more components are added. In general, models that are not methodologically data-driven suffer from feature dimension explosion or one-vs-all scaling issues when modeling multimodal data.

Recently, with the advent of deep learning and its applications in temporal data, RNNs have emerged as a strong phenomenological sequence-modeling framework \cite{SutskeverVinyalsEtAl2014, GravesMohamedEtAl2013}. Morton \etal analyzed the performance of RNNs for modeling driver behavior and demonstrated that Long Short-Term Memory (LSTM) networks can effectively model and predict vehicle motion distributions \cite{MortonWheelerEtAl2017}. They also showed that additional past history improves performance and that the models can perform well over longer time scales. In terms of handling multimodal data, the problem of HRI when the multimodality of future outcomes critically affects decision making has been studied \cite{SchmerlingLeungEtAl2018}. Schmerling \etal model human response dynamics to an autonomous vehicle's future actions using a Conditional Variational Autoencoder (CVAE) with LSTM subcomponents. Such a model allows for the efficient sampling of multimodal trajectories conditioned on future agent actions. As a result, we decided to base our multimodal, future-conditional trajectory prediction framework on a CVAE-LSTM architecture, similar to the one presented in \cite{SchmerlingLeungEtAl2018}. Our work differs from \cite{SchmerlingLeungEtAl2018} in that our approach can handle a variable number of agents, whereas theirs can only model one. A more detailed explanation of our approach follows in Section \ref{solution}.

Given this design choice, how does one efficiently model $N$ humans? Should existing LSTM architectures be scaled up to handle many input sizes or should models be developed which inherently consider many entities? The following subsections dive deeper into approaches from both possibilities.

\subsection{Scaling Existing Methods to Handle $N$ Humans} A common theme in works that scale existing methods to handle $N$ humans is the idea of replicating a base model to many humans and developing some kind of ``spatial interaction pooling layer." This layer combines the hidden states of nearby humans in order to condition a modeled human's hidden state with information from others. The key idea here is that an LSTM's hidden state should encode all information necessary to model future predictions, so having them from other humans implies having knowledge of what other humans are thinking.

Two recent approaches that exemplify this are the Social LSTM \cite{AlahiGoelEtAl2016} and Neural Physics Engine (NPE) \cite{ChangUllmanEtAl2017}. In \cite{AlahiGoelEtAl2016}, Alahi \etal study the problem of predicting human trajectories while taking into account the trajectories of other humans in the scene. They model human trajectories as outputs of an LSTM network. In order to scale up the usual LSTM architecture for $N$-human trajectory modeling, they introduce a ``Social Pooling" layer which pools together spatially close humans' hidden states and enables humans to generate trajectories that are mindful of others. 
They also show that interaction-informed LSTMs outperform prior $N$-human trajectory prediction frameworks such as GPs and the popular Social Forces model \cite{HelbingMolnar1995}.
The Neural Physics Engine (NPE) is a similar work that explores modeling physical object interactions (\eg billiard balls colliding) via explicit spatial pooling of $N$ bodies' latent states \cite{ChangUllmanEtAl2017}. It is very similar to the Social LSTM, even using the same underlying model and introducing a similar pairwise interaction ``pooling" layer. A key difference is that NPE only predicts an object's behavior one time step ahead per inference. They argue that such simple predictions hold accurately since they are not dealing with forces that operate at a range (\eg gravity, magnetism), but rather impulse dynamics.

\subsection{Inherently Modeling $N$ Humans' Interactions} Works in this category usually model humans and their interactions as nodes and edges in a probabilistic graphical model (PGM), with extensions that combine PGMs and deep learning. In \cite{WheelerKochenderfer2016}, Wheeler and Kochenderfer showed that PGMs can model the kinds of interactions we are interested in by generating novel, statistically-realistic highway scenes by modeling vehicles and their interactions as nodes and pairwise edges in a factor graph. They model the conditional distribution of a vehicle's state, as influenced by surrounding vehicles, as a Conditional Random Field (CRF), which is a form of undirected PGM. However, since running time is not an important factor for their posed problem, they rely on the slow Metropolis-Hastings algorithm for sampling (leading to long inference times). A similar modeling-and-generation framework was proposed in \cite{FouheyZitnick2014}, where a CRF model is presented that generates images by predicting object dynamics in scenes, taking into account the influence of other objects in the scene. They similarly model objects as nodes and their interactions as pairwise edges. Fouhey and Zitnick also verify that the model handles multimodality and show that the distribution of outputs is affected by conditioning on different input scene layouts. A limitation is that the paper does not address predictions across time, opting to produce a single image some time in the future rather than a set of images showing how the scene evolves to such a final state. Although Fouhey and Zitnick argue their model is spatiotemporal and provide absolute and relative velocities as features (trying to enforce the consistency of related objects), it is not as strong as explicitly adding temporal factors, an approach taken by \cite{JainZamirEtAl2016}. In addition to incorporating temporal factors, \cite{JainZamirEtAl2016} introduces a spatiotemporal model derived from a temporal CRF formulation, filling in the temporal gaps of prior work \cite{WheelerKochenderfer2016, FouheyZitnick2014}. Nodes model object behavior, spatial edges model spatial interactions, and temporal edges (between nodes of the same entity) model temporal interactions (\ie object trajectories across time).

Due to their inherent ability to model $N$ entities and mature surrounding body of work, CRFs are a natural choice for modeling the kinds of conditional interactions we are interested in, \ie the behavioral characteristics of other humans given an agent's future plan. Further, prior work has shown that treating spatiotemporal relations as edges in a graph and modeling them with deep learning techniques leads to strong performance \cite{JainZamirEtAl2016, DuWangEtAl2015, ByeonBreuelEtAl2015}. Accordingly, we will model each human in our problem as a graph node and pairwise relationships between nearby humans as a graph edge, modeling each component with deep learning. A more detailed explanation of our proposed model follows in Section \ref{solution}.

\section{PROBLEM FORMULATION}\label{problem_formulation}

\subsubsection*{Notation} We use superscripts, \eg $x^{(t)}$, to denote the values of quantities at discrete time steps, and the notation $x^{(t_1 : t_2)} = (x^{(t_1)}, x^{(t_1 + 1)}, \dots, x^{(t_2)})$ to denote a sequence of values at time steps in $[t_1, t_2]$ for $t_1 \leq t_2$.

\subsection{Interaction Dynamics}
Let the deterministic, time-invariant, discrete-time state space dynamics of $N$ humans and a future-conditioning agent be given by
\begin{equation} \label{dynamics}
x_{h_i}^{(t+1)} = f_{h_i}\left(x_{h_i}^{(t)}, u_{h_i}^{(t)}\right), \ \ x_{a}^{(t+1)} = f_{a}\left(x_{a}^{(t)}, u_{a}^{(t)}\right)
\end{equation}
respectively, where $x_{h_i}^{(t)} \in \mathbb{R}^{d_{x_h}}, x_a^{(t)} \in \mathbb{R}^{d_{x_a}}$ denote the states of human $i$ and the agent and $u_{h_i}^{(t)} \in \mathbb{R}^{d_{u_h}}, u_a^{(t)} \in \mathbb{R}^{d_{u_a}}$ denote their chosen control actions at time $t \in \mathbb{N}_{\geq 0}$. Let $x^{(t)} = (x_{h_1}^{(t)}, \dots, x_{h_N}^{(t)}, x_a^{(t)})$ and $u^{(t)} = (u_{h_1}^{(t)}, \dots, u_{h_N}^{(t)}, u_a^{(t)})$ denote the joint state and control of the $N+1$ entities.

Interactions end when the joint state first reaches a terminal set $\mathcal{T} \subset \mathbb{R}^{Nd_{x_h} + d_{x_a}}$ and let $T$ denote the final time step, $x^{(T)} \in \mathcal{T}$. For each $t < T$, we assume that all humans' next actions $u_{h_1, \dots, h_N}^{(t+1)}$ are drawn from a distribution conditioned on the joint interaction history $(x^{(0:t)}, u^{(0:t)})$ and the agent's next action $u_a^{(t+1)}$. Thus, $U_{h_1, \dots, h_N}^{(t+1)} \sim P( x^{(0:t)}, u^{(0:t)}, u_a^{(t+1)} )$, with drawn values denoted in lowercase. We further suppose that $U_{h_1, \dots, h_N}^{(t+1)}$ is distributed according to a probability density function (pdf) which we denote as $p(u_{h_1, \dots, h_N}^{(t+1)} | x^{(0:t)}, u^{(0:t)}, u_a^{(t+1)})$. Finally, we also assume that all past states and actions by all entities are fully observable. This means that an agent may reason about $U_{h_1, \dots, h_N}^{(t+1: t+S)}$, representing all humans' response sequences to agent actions $u_a^{(t+1 : t+S)}$ over a horizon of length $S$, by propagating \eqref{dynamics} and sampling $U_{h_1, \dots, h_N}^{(t+1)}$.

\subsection{Specific Scenario}
Although we have developed our approach generally for multiple human-agent and human-human interactions, we chose basketball playing as the scenario to present our model's performance because the setting is challenging, highly dimensional, and rich with multimodal, multi-human interactions. Further, high-quality trajectory data is freely available from prior industry-led data-gathering campaigns.

We chose this scenario with the understanding that we can apply our methodology to robotics applications such as autonomous driving. This understanding is born from the realization that human trajectory modeling and applications such as human driver modeling are specific examples of the more general problem of modeling node behavior and edge interactions in a general undirected graph. Further, applications such as driver and athlete modeling are quite similar in that they are both inherently rife with multimodality, require accounting for multi-human interactions, are continuous prediction problems, and are difficult to model without a data-driven approach.

It is important to note that we purposely do not model the basketball in this work. Although encoding specific domain knowledge like the ball's position would probably help with modeling performance, we exclude it to keep our model general since many real-world scenarios do not have a singular goal object, \eg human driving behavior. %

\begin{figure}[t]
  \centering
  \includegraphics[width=0.90\linewidth]{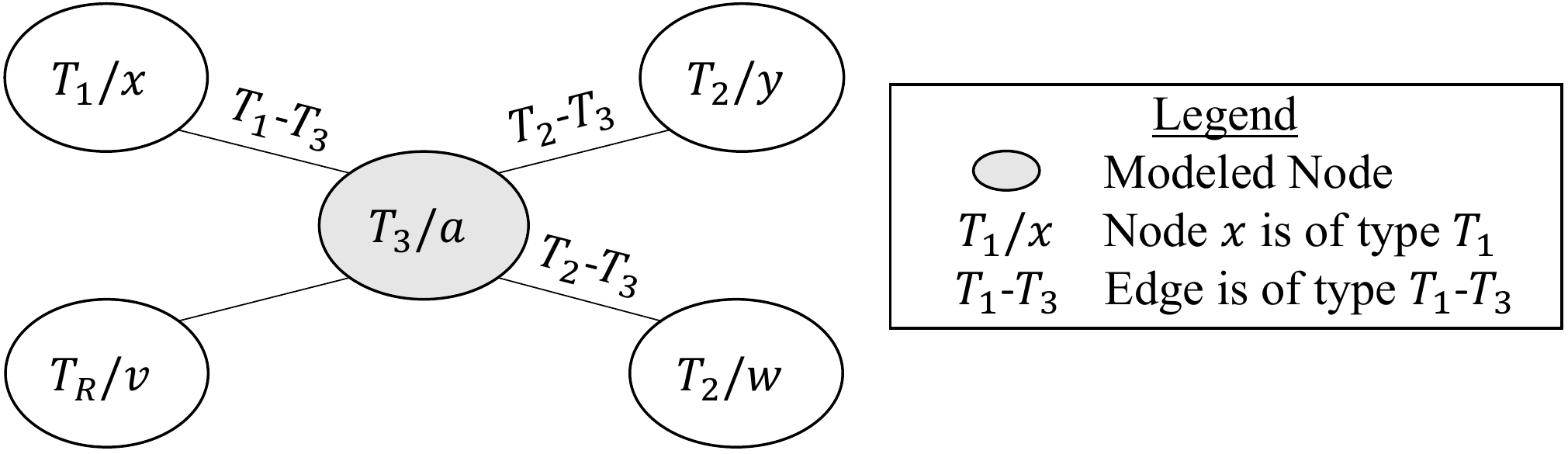}
  
  \vspace{0.2cm}
  
  \includegraphics[width=0.90\linewidth]{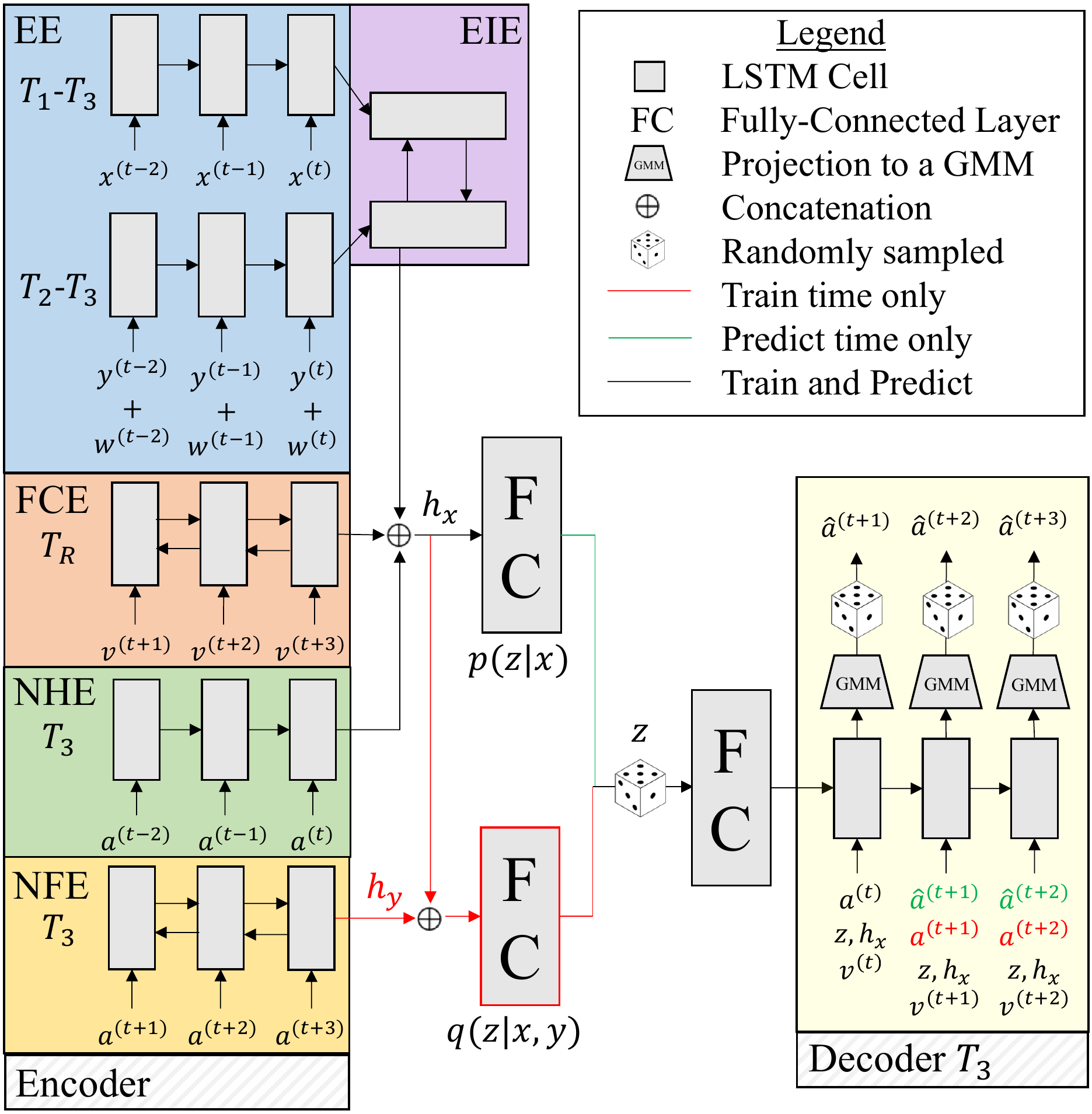}
  \caption{\textbf{Top}: An example graph with five nodes. $a$ is our modeled node and is of type $T_3$. It has four neighbors: $y, w$ of type $T_2$; $x$ of type $T_1$; and the agent $v$. \textbf{Bottom}: Our corresponding architecture for node $a$. The EE and EIE (``Edge Encoder" and ``Edge Influence Encoder") are responsible for encoding the effects neighbors have on the modeled node. The FCE (``Future Conditional Encoder") models the high-level semantics of a candidate agent future. The NHE (``Node History Encoder") represents the modeled node's past. The NFE (``Node Future Encoder") is only used during training and helps shape our latent representation. The Decoder samples a latent state $z$ and generates trajectories for the modeled node. Inputs to the model are concatenated states ($x$) and actions ($u$). This figure is best viewed in color.}
  \label{fig:arch_example}
  \label{fig:architecture}
\end{figure}

\subsubsection*{Basketball Playing}
Let $(l, w)$ be the coordinate system for a basketball court where $l$ denotes longitudinal distance along the length of the court (with 0 at the leftmost basket and positive values to the right) and $w$ denotes lateral court position (with 0 at the bottom-most extent of the court and positive values above) when viewed from above with the home team's logo upright. We consider the center of mass of human $i$
to obey single-integrator dynamics (\ie $u_{h_i} = [\dot{l}_{h_i}, \dot{w}_{h_i}], x_{h_i} = [l_{h_i}, w_{h_i}]$ in continuous form). This is an intuitive choice as an athlete's movements are all position-changing, \eg walking increases one's position along a direction, running does so faster. We enforce an upper bound of 12.42m/s on any human's speed, which is the current footspeed world record set by Usain Bolt \cite{GraubnerBuckwitzEtAl2009}.

\section{$N$ HUMAN MODELING}\label{solution}
We propose a solution combining CVAEs, LSTM networks, and a relationship-encoding graphical structure to produce high-quality multimodal trajectories that models and predicts the future behaviors of $N$ humans. Our full architecture is illustrated in Fig. \ref{fig:architecture}.

Considering a fixed prediction time step $t$, let $\mathbf{x} = (x^{(0:t)}, u^{(0:t)}, u_a^{(t+1:t+S)})$ be the input to the model (neighboring human interaction history and candidate agent future) and $\mathbf{y} = u_h^{(t+1:t+S)}$ be the desired prediction output (all humans' futures). We wish to learn a pdf $p(\mathbf{y} | \mathbf{x})$. In using a CVAE to do this, we introduce a latent variable $\mathbf{z}$ so that $p(\mathbf{y} | \mathbf{x}) = \int_{\mathbf{z}} p(\mathbf{y} | \mathbf{x}, \mathbf{z})p(\mathbf{z} | \mathbf{x}) \text{d}\mathbf{z}$. $\mathbf{z}$'s purpose is to model latent structure in the interaction which may both improve learning performance and enable interpretation of results \cite{SohnLeeEtAl2015}. In our work, $p(\mathbf{y} | \mathbf{x}, \mathbf{z})$ and $p(\mathbf{z} | \mathbf{x})$ are modeled using neural networks that are fit to maximize the likelihood of a dataset $\mathcal{D} = \{(\mathbf{x}_i, \mathbf{y}_i)\}_i$ of observed interactions. This optimization is performed by instead maximizing the evidence-based lower bound of the log-likelihood $\log p(\mathbf{y} | \mathbf{x})$ by importance sampling $\mathbf{z}$ from a neural network approximation $q(\mathbf{z} | \mathbf{x}, \mathbf{y})$ of the true posterior $p(\mathbf{z} | \mathbf{x}, \mathbf{y})$ \cite{Doersch2016}.

An LSTM network is a specific type of RNN where each node contains internal gating functions that enable a finer control of information flow \cite{HochreiterSchmidhuber1997}. Each node's internal computations are as follows \cite{KalchbrennerDanihelkaEtAl2015}:
\begin{equation}\label{lstm}
    \begin{split}
        i_t &= \sigma(W^{(i)} x_{t} + U^{(i)} h_{(t-1)} + b^{(i)}) \\
        f_t &= \sigma(W^{(f)} x_{t} + U^{(f)} h_{(t-1)} + b^{(f)}) \\
        o_t &= \sigma(W^{(o)} x_{t} + U^{(o)} h_{(t-1)} + b^{(o)}) \\
		\Tilde{c}_t &= \tanh(W^{(c)} x_{t} + U^{(c)} h_{(t-1)} + b^{(c)}) \\
		c_t &= f_t \odot c_{(t-1)} + i_t \odot \Tilde{c}_t\\
		h_t &= o_t \odot \tanh(c_t)
    \end{split}
\end{equation}
where $W^{(i)}, W^{(f)}, W^{(o)}, W^{(c)}$ are recurrent weight matrices; $U^{(i)}, U^{(f)}, U^{(o)}, U^{(c)}$ are input projection matrices; $b^{(i)}, b^{(f)}, b^{(o)}, b^{(c)}$ are bias terms; and $\odot$ denotes element-wise multiplication. $h_t$ is commonly referred to as an LSTM's ``hidden state" and $c_t$ its ``memory vector", partially responsible for controlling state updates and outputs. $h_t$ is also the output at each time step.

We proceed by describing our model and highlighting key components that enable it to efficiently scale to modeling $N$ humans' behaviors.

\subsubsection*{Graphical Representation}
When presented with an input problem, we first automatically create an undirected graph representing the scene (Fig. \ref{fig:scenarios} shows an example of this). This enables our model to be applied to different problem domains as graphs are abstract representations of entities and their relationships in a scene, free of domain-specific information. In this work, we form edges based on entities' spatial proximity.

\subsubsection*{Encoding Influence from Neighboring Humans}
We use Edge Encoders (EEs) to incorporate influence from neighboring nodes. Each encoder is an LSTM network with 8 hidden dimensions and standard sigmoid/tanh nonlinearities. Note that we focus on edge types rather than individual edges. This allows for more efficient scaling and dataset efficiency as we reuse edge encoder weights across edges of the same type. We combine the states of all neighboring nodes of a specific edge type by summing them and feeding the result into the appropriate edge encoder, obtaining an edge influence representation. This representation is then merged with other edge influences via an Edge Influence Encoder (EIE) to obtain a total edge influence encoding. The EIE is a bi-directional LSTM with 8 hidden dimensions and standard sigmoid/tanh nonlinearities. We feed encoded edges as inputs to the EIE and output the final concatenated forward and backward hidden and memory vectors. We choose to combine representations in this manner rather than via concatenation in order to handle a variable number of neighboring nodes with a fixed architecture. We opted for a bi-directional architecture since we wish to be able to encode the same total edge influence regardless of ordering. Experiments were performed to select the highest performing edge input combination method and edge influence combination method, their details and results are shown in Section \ref{experiments}.

\subsubsection*{Encoding Node History}
We use a Node History Encoder (NHE) to represent a node's state history. It is an LSTM network with 32 hidden dimensions and standard nonlinearities.

\subsubsection*{Encoding Candidate Agent Futures}
We use a Future Conditional Encoder (FCE) to represent a neighboring future-conditional agent's effect on us. It is a bi-directional LSTM with 32 hidden dimensions and standard sigmoid/tanh nonlinearities. We opted to use a bi-directional LSTM in order to capture higher-level semantics in the candidate agent future, allowing the overall character of our generated trajectory to be affected rather than merely generating a short-term reaction. The effect of different candidate agent futures is shown in Section \ref{experiments}.

\subsubsection*{Generating Trajectories}
For our decoder, we use an LSTM network with 128 hidden dimensions and standard sigmoid/tanh nonlinearities whose outputs are Gaussian Mixture Model (GMM) parameters with $N_{GMM}$ components. These GMMs are over the human action space. We then sample from these GMMs to output trajectories. 

\subsubsection*{Scaling via Weight Sharing}\label{weight_sharing}
We handle $N$ humans by replicating the model in Fig. \ref{fig:architecture} for each node. Although this causes computation and memory usage to scale linearly with respect to the number of nodes, it is necessary in order to generate individual future trajectories. In practice, this does not cause the number of parameters to scale linearly as we implement extensive weight sharing across the model, meaning the number of parameters our method uses scales proportional to the number of node and edge \textit{types} rather than the number of individual nodes or edges. We share weights across every part of the model: EEs of the same edge type share weights; EIEs, NHEs, and NFEs of the same node type share weights; and the FCE shares its weights across all models. Most importantly, this means that our model is very dataset efficient, with a single data point able to update parts of the model multiple times (\eg a graph with multiple instances of a node type).

\subsubsection*{Additional Considerations}
We enforce the notion that agents only affect their immediate neighbors, and that any resulting effects will ripple out to their neighbors' neighbors, and so on. If one believed that the agent affects all others regardless of spatial distance, there would be an FCE for every human in the scene. Since we wish to enforce this locality notion, for humans not connected to the agent we zero-out the FCE from their CVAE-LSTM architecture. This has the added benefit of reducing overall model complexity since a majority of humans will be far away from the agent (assuming the agent has less than $N/2$ edges). Note that this means we also explicitly model human-human interactions, rather than only agent-human ones. This extra work might be viewed as unnecessary and that modeling all of the extra human-human edges needlessly adds computation time. However, we argue that considering human-human interactions in addition to human-agent interactions is indeed necessary, as it allows the agent to reason more strongly about potential human behaviors. For example, take a three-lane driving scenario where an autonomous vehicle in the left lane is predicting surrounding human trajectories, with humans next to the vehicle in the middle and right lanes. It goes without saying that a driver cannot turn into a lane with a vehicle in it next to them. Without acknowledging this human-human interaction, an agent may reason that the human can indeed change lanes and proceed to turn into the human's lane, creating a dangerous situation. Generalizing, the ability to better model human behaviors (taking into account their own interactions and resulting restrictions) leads to better predictions and better-informed agent trajectories.

Although we base our graphical model on the structure of CRFs, it is important to note that our graphical model is not probabilistic and is not meant to model the joint distribution over the outputs. We instead focus on learning the dependencies between the outputs via structural sharing of LSTM networks between the outputs, similar to \cite{JainZamirEtAl2016}.

To sample from our model, we execute forward passes of our architecture for each node. This is different from the way one would sample from a CRF, \eg using a method like Metropolis-Hastings \cite{Hastings1970}. We choose not to sample in this manner as it is too slow for robotic use cases. In practice, one would replan with fast update rates and encode information in a stateful manner (\ie with LSTM hidden states), so we make a reasonable assumption that the information encoded in previous time steps' hidden states is globally informed and sufficient for decision-making about the present and future.

\section{EXPERIMENTS}\label{experiments}

All of our source code, trained models, and data can be found at \url{https://github.com/StanfordASL/NHumanModeling}. Training and experimentation was performed on a single Amazon Web Services (AWS) p2.8xlarge EC2 Instance with TensorFlow \cite{Abadi2015}. Scalability profiling was performed on multiple AWS m4.4xlarge EC2 instances. Throughout this section, we use a model's negative log-likelihood (NLL) prediction loss on the validation set as an evaluation metric, meaning that lower values are better. As discussed in Section \ref{weight_sharing}, our model is very dataset efficient. In fact, all of the results in this paper were obtained after 230 training steps, orders of magnitude less steps than typical deep learning models.

\subsection{Dataset}
A collection of 97 ``plays" recorded from 26 distinct professional basketball players during the 2015-16 NBA season \cite{Linou2016}. In prior seasons, the NBA installed SportVU Player Tracking technology in every arena. The technology captures data every 40ms, tracking positions of each of the 10 players during a game. This equates to $T \approx 600$ assuming a 24s play duration, which is the length of the shot clock during a single possession. We define a ``play" as any contiguous amount of time-reducing game play (\ie we ignore breaks, free throws, fouls, etc.). Taking a prediction horizon of $S = 15$ (600ms into the future), the total dataset contains approximately 51,000 $\mathbf x = (x^{(0:t)}, u^{(0:t)}, u_a^{(t+1 : t+S)}), \mathbf y = u_h^{(t+1 : t+S)}$ data points. We chose this time horizon because a professional basketball player can cover distances of $\sim$4.5m in 600ms \cite{ios_NBA:2017}. We set aside 10 plays from an NBA game not present in the training dataset to form our validation set with which we evaluate and visualize our model's performance.

\subsection{Importance of Modeling Multimodality}
Our explicit modeling of multimodality is a crucial part of our model's performance.
To verify this, we evaluated our model's performance after reducing the number and dimension of latent variables in the CVAE and the number of GMM components it uses. The results are summarized in Table \ref{table:multimodality_perf}.

We begin by limiting the model to only having one GMM output component and using one discrete latent variable ($N_K = K_z = N_{GMM} = 1$). Without the explicit ability to handle multimodality, a validation NLL of $554.24$ is obtained, compared to our original model's value of $-19.00$. Note that this reduced version of our model is representative of current state-of-the-art approaches such as the Structural-RNN \cite{JainZamirEtAl2016} and Social LSTM \cite{AlahiGoelEtAl2016} since they use LSTMs for node behavior modeling and do not inherently account for multimodality, relying instead on edge representations to condition outputs. This variation of our model is also representative of the performance that IRL and standard GPR would obtain, being non-multimodal approaches themselves. As can be seen, explicitly handling multimodality is crucial for performance, and leads to our model outperforming current state-of-the-art approaches.

From here, we can increase the expressiveness of the baseline by increasing the number of GMM components and latent variables. Increasing only the number of GMM components ($N_{GMM} = 16$) yields a validation NLL of $-12.30$. Increasing only the number of latent variables ($N_K = 2, K_z = 5$) yields a validation NLL of $263.67$. As expected, our full model significantly outperforms restricted versions.

\vspace{-0.3cm}
\begin{table}[htbp]
\caption{Model Performance across CVAE / GMM Parameter Choices}
\vspace{-0.25cm}
\centering
\begin{tabular}{|c|c|}
\hline
\textbf{CVAE / GMM Parameters} & \textbf{Validation NLL} \\
\hline
$N_K = 1, K_z =1,  N_{GMM} = 1$ & $554.24$\\

$N_K = 2, K_z = 5, N_{GMM} = 1$ & $263.67$\\

$N_K = 1, K_z = 1, N_{GMM} = 16$ & $-12.30$\\

$N_K = 2, K_z = 5, N_{GMM} = 16^{\mathrm{a}}$ & $-19.00$\\
\hline
\multicolumn{2}{l}{$^{\mathrm{a}}$These are our full model's parameters.}
\end{tabular}
\label{table:multimodality_perf}
\end{table}
\vspace{-0.5cm}

\subsection{Edge Radii}
Table \ref{table:edge_radii}
shows our model's validation performance with respect to different edge radii. We trained one model per edge radius and evaluated its performance on our validation set. We found that as radius increases, performance also increases. This matches intuition as a larger edge radius takes into account more surrounding behavior, leading to a more comprehensive picture of the scene around the modeled node. While it may be tempting to set the edge radius high for the best performance, there would be an associated penalty with respect to larger memory usage, computation time, and number of parameters. We discuss this scaling in subsection \ref{expt:scalability}. One way to practically choose an edge radius for a given problem would be to visualize the Pareto front that arises when varying edge radius and measuring the resulting model's memory usage, computation time, and number of parameters, selecting what best suits hardware or time constraints.

\vspace{-0.3cm}
\begin{table}[htbp]
\caption{Model Performance across Edge Radii}
\vspace{-0.25cm}
\centering
\begin{tabular}{|c|c|}
\hline
\textbf{Edge Radius} & \textbf{Validation NLL} \\
\hline
$1$m & $-19.00$\\

$2$m & $-21.94$\\

$3$m & $-22.17$\\

$4$m & $-23.53$\\

$5$m & $-24.73$\\
\hline
\end{tabular}
\label{table:edge_radii}
\end{table}
\vspace{-0.5cm}

\subsection{Edge Encoder Input Combination Method}
To select the best edge input combination method, we trained multiple models that use different edge input methods and evaluated their validation performance. The first model sums inputs to the EEs, the second averages inputs. We obtained a validation NLL of $-23.23$ for the sum method and $-21.94$ for the mean method. It is intuitive that the sum method performs better because this method retains information about how many inputs were originally present, whereas averaging does not.

In an effort to reduce information loss, one may desire to have a model with an EE per neighbor. We found that this approach is prohibitively expensive. Using such a scheme for a small graph with 10 nodes and 10 edges resulted in a model requiring 243 MB more memory (701 MB vs. 458 MB) and suffering a 29x slower average runtime as measured over 100 forward passes (7.31s vs. 0.25s per forward pass) when compared to our architecture over the same graph.

\subsection{Edge Influence Encoder Architecture}
To choose the best EIE architecture, we trained multiple models with different EIE architectures and compared their validation performance. ``Sum" means we sum all of the encoded edge vectors, ``Max" means we perform an element-wise maximum over the encoded edge vectors, and ``Bi-LSTM" means we construct a bi-directional LSTM network whose inputs are the encoded edge vectors and output is the final concatenated forward and backward hidden and memory vectors. We then feed this combined representation into the rest of the architecture. We found that a bi-directional LSTM architecture performs the best (with a validation NLL of $-21.94$), which makes sense as the network can learn the best combination function rather than only being a simple element-wise sum (validation NLL of $-15.79$) or max (validation NLL of $-21.04$).

\subsection{Scalability}\label{expt:scalability}
\begin{figure*}[ht]
  \centering
  \begin{minipage}{0.68\linewidth}
    \includegraphics[width=\linewidth]{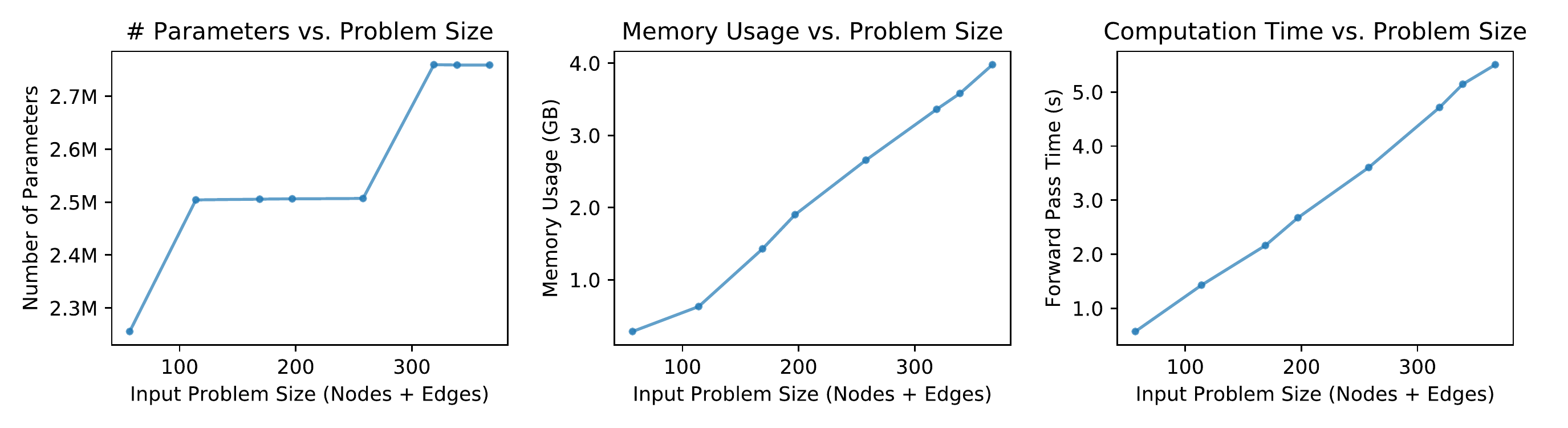}
  \end{minipage}
  \begin{minipage}{0.31\linewidth}
    \includegraphics[width=\linewidth]{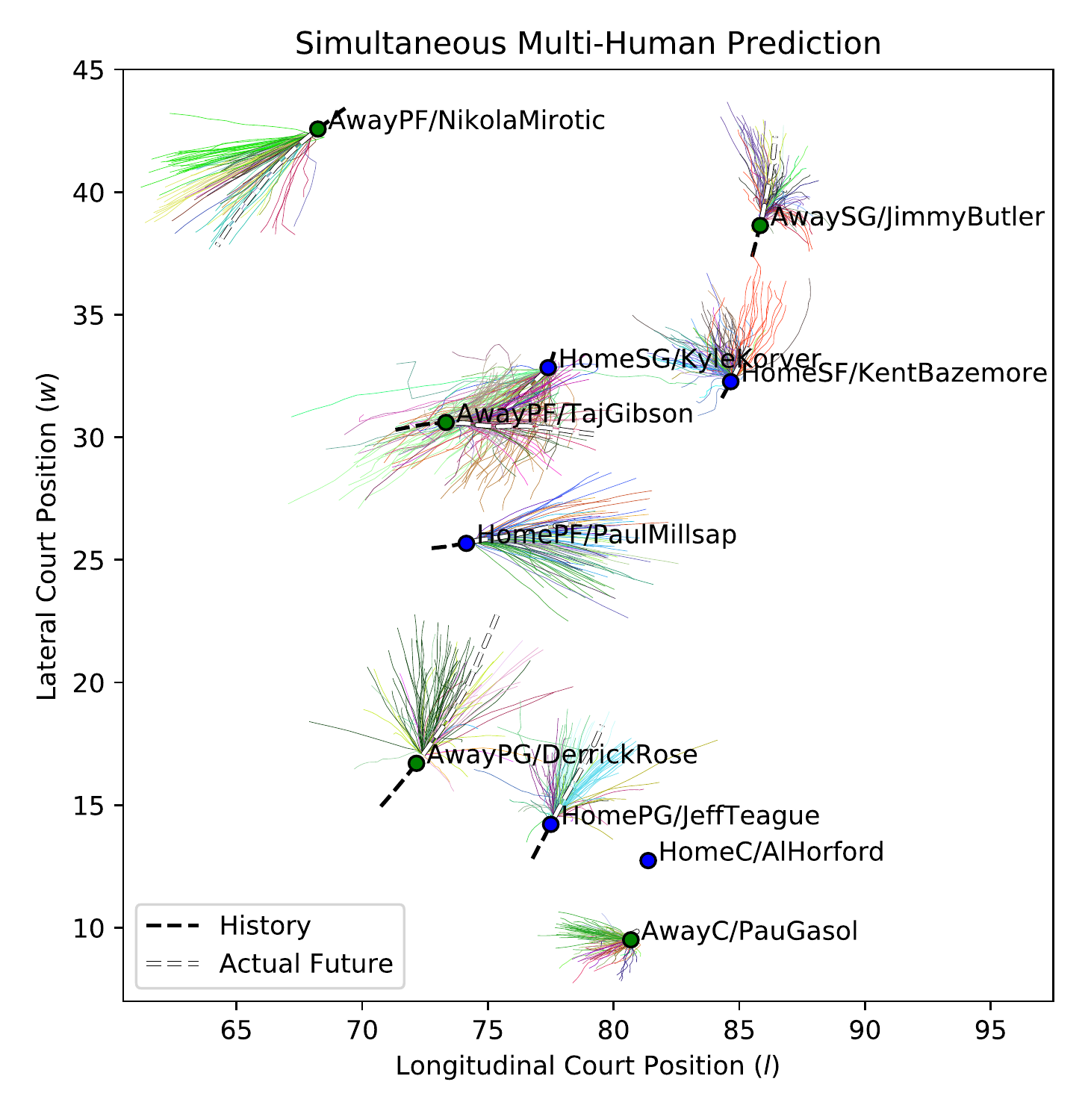}
  \end{minipage}
  \caption{
  \textbf{Leftmost}: How the number of model parameters scales with the size of the problem. The jumps are caused by a new set of players being encountered in the data, causing our model to increase in size. \textbf{Middle Left}: Model memory usage with respect to problem size. We expect this to be linear as each additional node and edge directly adds LSTM networks to our model. \textbf{Middle Right}: Model forward computation time with respect to problem size. We expect this to scale linearly with problem size following the same reasoning as with memory usage. \textbf{Rightmost}: Our model's simultaneous prediction of all nodes' potential futures in a scene. 100 samples are drawn per node, except for Al Horford whose future we condition on. No history or future are visible for Horford because he is standing still. Colored lines are sampled predictions where color indicates the value of the latent variable $z$.}
  \label{fig:scalability}
  \label{fig:whole_court}
\end{figure*}

Fig. \ref{fig:scalability} shows how our model scales with respect to input problem size (number of nodes and edges). Scaling performance was obtained by profiling a copy of our model running as the lone process on a m4.4xlarge AWS EC2 instance. As can be seen, our model's computation time and memory usage scales linearly with respect to problem size. This makes sense as additional nodes and edges in a graph directly increase the size of our model (\ie by adding architectures to model them). Note that the number of parameters stays constant for problems with the same number of unique node and edge types, this is due to our aggressive weight sharing scheme (discussed in Section \ref{solution}). This scaling behavior is another way that our approach outperforms methods such as GPR, which have very poor scaling performance with respect to input dataset and problem size \cite{RasmussenWilliams2006}.

\subsection{Analysis of Model Properties}
Taking into account results from the previous subsections, we trained a version of our model with a 2m edge radius, sum edge input method, and bi-directional LSTM edge influence method (exactly as depicted in Fig. \ref{fig:architecture}). We then plot several of its future predictions on the validation set in different interesting scenarios. In them, we can see the model exhibiting desired behaviors from our design choices. In the figures, each player's name and type are printed next to the node that represents them in the format ``PlayerType/PlayerName". A player's type is a concatenation of their team (``Home" or ``Away") and their role (``C", ``PF", ``SF", ``SG", or ``PG"). We used these player types as node types in our model, meaning we have at most 10 unique node types.

The ability of our model to simultaneously predict $N$ humans' behaviors can be seen in Fig. \ref{fig:whole_court} (rightmost), showing a full output for a scene. In it, trajectory samples can be seen for each node, colored by their respective $z$ value. We can see that different $z$ values lead to different plausible trajectory groups, showing that the model learns a good separation of high-level behaviors. 

\begin{figure}[t]
  \centering
  \includegraphics[width=0.46\linewidth]{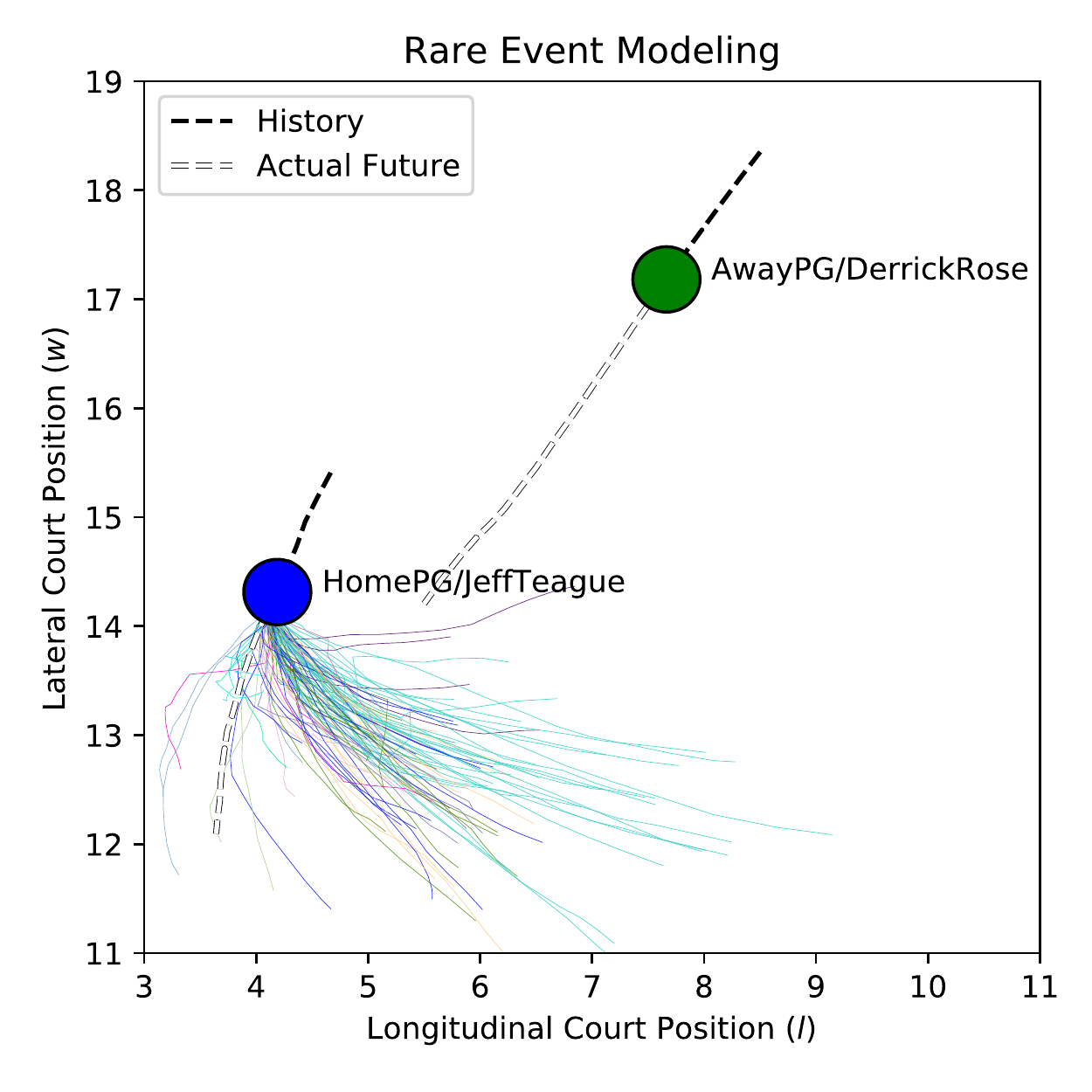}
  \includegraphics[width=0.52\linewidth]{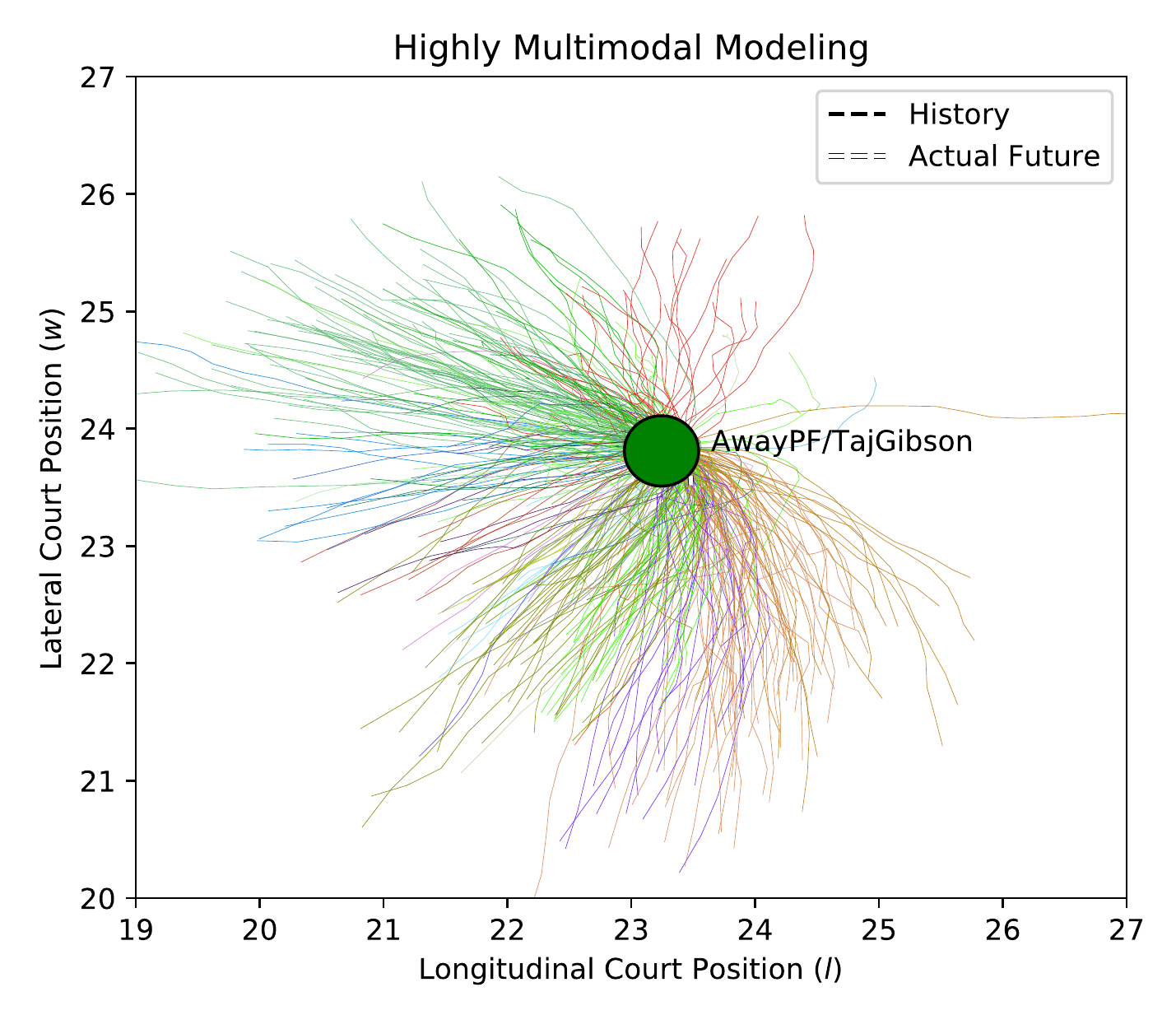}
  \caption{\textbf{Left}: A scenario where the majority of our 100 generated trajectories predict that Jeff Teague would cut across the front of Derrick Rose. However, Teague instead continues forward, a rare event. Since our model captures uncertainty, we have a prediction (in blue) that matches Teague's actual future. For clarity, we omitted our model's trajectory predictions for Derrick Rose. \textbf{Right}: A stationary player for which our model generates 400 different futures. This illustrates the highly multimodal nature of our model, and how it can generate many plausible high-level behaviors and corresponding low-level trajectories. Colored lines are sampled predictions where color indicates the value of the latent variable $z$.}
  \label{fig:rare_events}
  \label{fig:clear_multimodal}
\end{figure}

Fig. \ref{fig:rare_events} (left) depicts a scenario where a majority of our model's predicted trajectories do not match the actual player's future. However, due to our model's handling of uncertainty, it can still generate a prediction that matches the player's actual future. A model which only predicts the most likely future would have a difficult time capturing such events.

Even if a model can handle uncertainty, the ability to generate multiple different high-level behaviors is necessary to capture the potential for multiple different semantic behaviors. Fig. \ref{fig:clear_multimodal} (right) shows this ability in our model as it generates many clear, high-level behaviors visible via the colored trajectory groupings. These make sense as the modeled player is stationary and can realistically move in any direction in the future. Further, since we are modeling basketball players, there is a bias to move towards the opposing team's basket (which is to the left in this case). This is captured by our model as well, and is evidenced by the relative lack of trajectories originating from the top right side of the player.

\begin{figure}[t]
  \centering
  \includegraphics[width=0.65\linewidth]{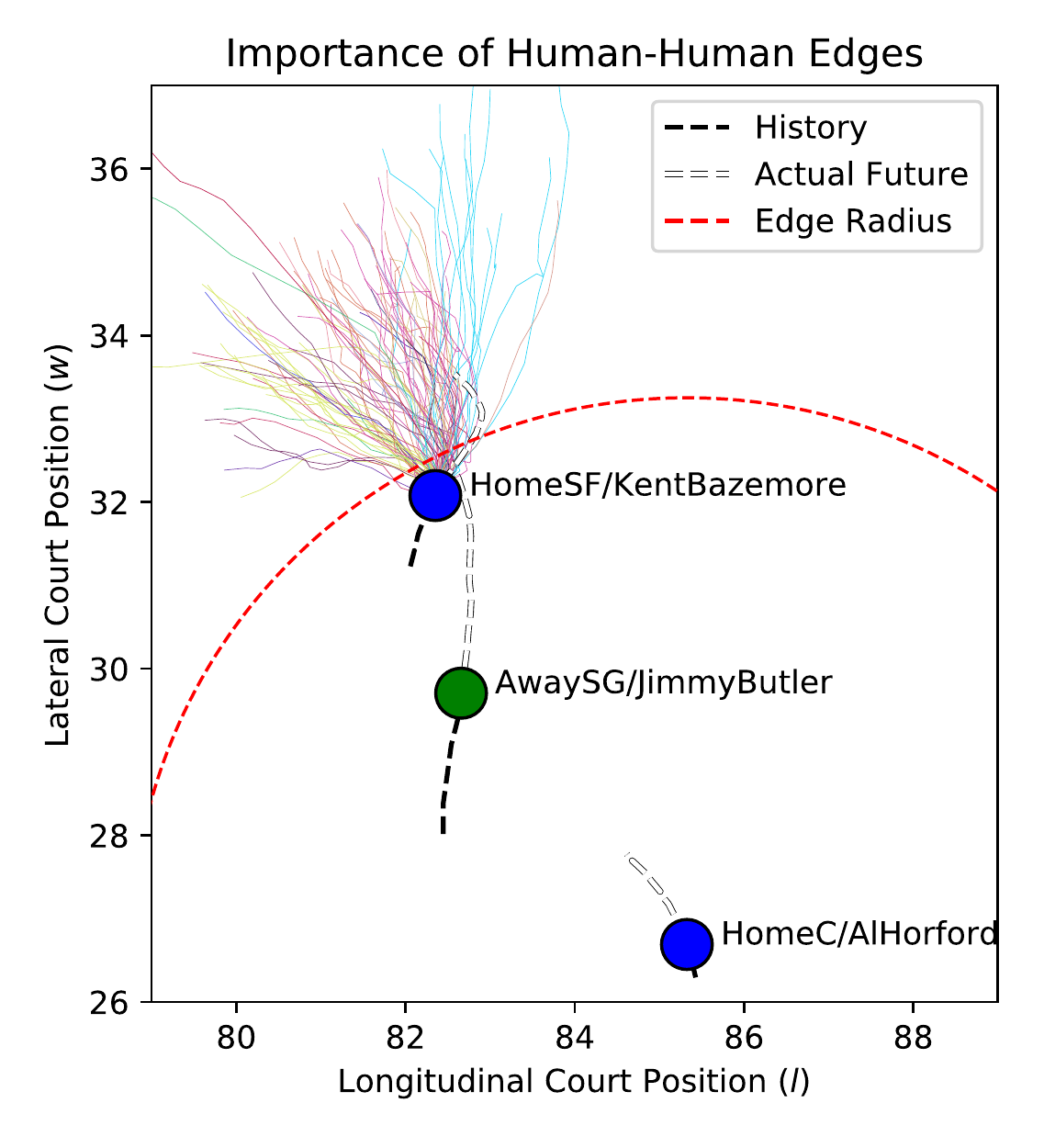}
  \includegraphics[width=0.33\linewidth]{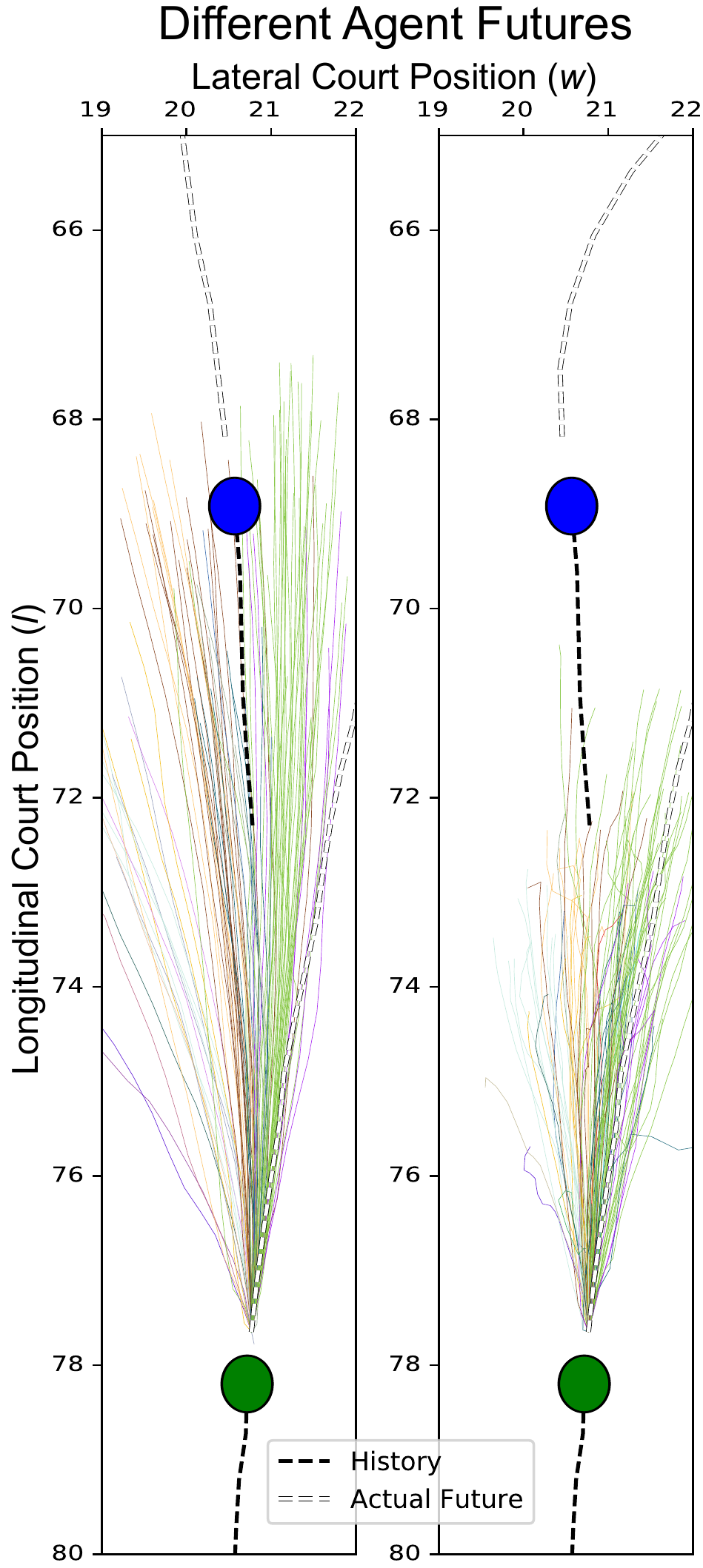}
  \caption{\textbf{Left}: An example where modeling human-human features is crucial. Without doing so, one might predict that Bazemore has options to move to the right, but this would place him closer to his opponent and less available to receive a pass or drive to score. Our model captures this restriction and better matches reality since Bazemore does actually move forward and cut left. The dashed red circle indicates the edge radius of the future-conditioning agent Al Horford. Butler's sampled futures are omitted for clarity. \textbf{Right}: A demonstration of the future-conditional aspect of our model, showing 100 sampled futures with respect to two candidate agent futures. Colored lines are sampled predictions where color indicates the value of the latent variable $z$.}
  \label{fig:human_human_importance}
  \label{fig:diff_futures}
\end{figure}

Even with the ability to model multimodality and uncertainty, one still needs to consider human-human interactions in addition to human-agent interactions. Fig. \ref{fig:human_human_importance} (left) illustrates the importance of modeling such edges. In the scenario depicted, none of our model's 100 sampled futures are on the right side of Kent Bazemore, since he's imminently affected by Jimmy Butler and would be less available to receive a pass or drive to score. A model that does not consider such interactions may ignore such restrictions on behavior.

Fig. \ref{fig:diff_futures} (right) shows that our model is future-conditional, as desired. Further, the sampled output actions make sense with respect to the candidate agent future as the first case allows Gibson to continue running straight whereas the second case forces Gibson to veer to the right to follow Horford.

\section{CONCLUSIONS}\label{conclusion}
Progress on this problem has the potential to significantly improve both the safety and performance of robotic systems. For example, autonomous vehicles surrounded by other cars behave far more conservatively than humans because they have few models of how humans will behave, potentially leading to confusion among human drivers. Our system would allow the vehicle to predict human reactions and take trajectories that are more natural and representative of driver expectations. Another example lies in collaborative tasks with humans such as packing boxes in a warehouse or preparing food. Our method would allow for the anticipation of human actions and improved efficiency in meeting them, \eg ``the robot moved its arms out of my way as I handed my sous chef a plate." 

In this work, we have presented a novel generative model for multimodal multi-human behavior that takes into account interaction and scales to real-world use-cases. Further, we have shown that this method is state-of-the-art in that it scales better than existing methods \cite{RasmussenWilliams2006} and generates outputs that are a better estimate of future human behavior compared to current state-of-the-art methods \cite{AlahiGoelEtAl2016, JainZamirEtAl2016}. Future directions include collecting data on real-world driving scenarios with multi-human interaction, applying this methodology to model human behavior, and incorporating the results into planning frameworks to better inform decision making.

\subsubsection*{Acknowledgment}
We thank Kevin Jia for his insights on basketball and considerations about what factors are important for modeling player trajectories.

\addtolength{\textheight}{-0.11cm}
\bibliography{ASL_papers,main}

\end{document}